\documentclass[journal]{IEEEtran}

\ifCLASSINFOpdf
  \usepackage[pdftex]{graphicx}
  \graphicspath{{./fig}}
  \DeclareGraphicsExtensions{.pdf,.jpeg,.png}
\else
  \usepackage[dvips]{graphicx}
  \graphicspath{{./fig/}}
  \DeclareGraphicsExtensions{.eps}
\fi

\usepackage{color}

\usepackage{amsmath}
\usepackage{amssymb}
\usepackage{amsthm}

\usepackage{algorithm}
\usepackage{algpseudocode}
\usepackage{array, makecell} %
\usepackage{booktabs}
\usepackage{multirow}

\begin{document}

\title{Evolutionary Architecture Search for Graph \\Neural Networks}
\author{}
\author{Min~Shi,
        David~A.~Wilson,~\IEEEmembership{Student~Member,~IEEE},
        Xingquan~Zhu,~\IEEEmembership{Senior~Member,~IEEE},
        Yu Huang,
        Yuan Zhuang,~\IEEEmembership{Member,~IEEE}, 
        Jianxun Liu,
        and~Yufei~Tang{*},~\IEEEmembership{Member,~IEEE}%
        \thanks{This work was supported in part by the U.S. National Science Foundation through Grant Nos. IIS-1763452, CNS-1828181, IIS-2027339, and OAC-2017597, and an Early-Career Research Fellowship from the Gulf Research Program (GRP) of the National Academies of Sciences, Engineering, and Medicine (NASEM).}
\thanks {M. Shi is with the Department of Computer \& Electrical Engineering and Computer Science, Florida Atlantic University, USA, and the School of Computer Science and Engineering, Hunan University of Science and Technology, Hunan, China. E-mail: toshimin132@gmail.com.}
\thanks {D. Wilson, Y. Huang, X. Zhu, and Y. Tang are with the Department of Computer \& Electrical Engineering and Computer Science, Florida Atlantic University, Boca Raton, FL 33431 USA. E-mails: \{davidwilson2016, yhwang2018@fau.edu, xzhu3, tangy\}@fau.edu.} \protect
\thanks {Yuan Zhuang is with the State Key Laboratory of Information Engineering in Surveying, Mapping and Remote Sensing, Wuhan University, 129 Luoyu Road, Wuhan 430079, China. E-mail: yuan.zhuang@whu.edu.cn.}
\thanks {Jianxun Liu is with the School of Computer Science and Engineering, Hunan University of Science and Technology, Hunan, China. E-mail: ljx529@gmail.com.}
\thanks{{*} Corresponding author.}
}

\markboth{}%
{Shi \MakeLowercase{\textit{et al.}}: Evolutionary Architecture Search for Graph Neural Networks}

\maketitle

\begin{abstract}
Automated machine learning (AutoML) has seen a resurgence in interest with the boom of deep learning over the past decade. In particular, Neural Architecture Search (NAS) has seen significant attention throughout the AutoML research community, and has pushed forward the state-of-the-art in a number of neural models to address grid-like data such as texts and images. However, very litter work has been done about Graph Neural Networks (GNN) learning on unstructured network data. Given the huge number of choices and combinations of components such as aggregator and activation function, determining the suitable GNN structure for a specific problem normally necessitates tremendous expert knowledge and laborious trails. In addition, the slight variation of hyper parameters such as learning rate and dropout rate could dramatically hurt the learning capacity of GNN. In this paper, we propose a novel AutoML framework through the evolution of individual models in a large GNN architecture space involving both neural structures and learning parameters. Instead of optimizing only the model structures with fixed parameter settings as existing work, an alternating evolution process is performed between GNN structures and learning parameters to dynamically find the best fit of each other. To the best of our knowledge, this is the first work to introduce and evaluate evolutionary architecture search for GNN models. Experiments and validations demonstrate that evolutionary NAS is capable of matching existing state-of-the-art reinforcement learning approaches for both the semi-supervised transductive and inductive node representation learning and classification. 
\end{abstract}


\IEEEpeerreviewmaketitle

\section{Introduction} \label{sec:intro}

\newcommand{\nickname}{Genetic-GNN}
\newcommand{\structset}{\textbf{S}}
\newcommand{\paramset}{\textbf{P}}
\newcommand{\numiter}{I}
\newcommand{\graph}{G}
\newcommand{\nodeset}{\textbf{V}}
\newcommand{\edgeset}{\textbf{E}}
\newcommand{\featset}{\textbf{X}}
\newcommand{\spacer}{\mathbb{R}}
\newcommand{\featnum}{n_f}
\newcommand{\embedh}{\textbf{h}}
\newcommand{\embedsize}{n_d}
\newcommand{\hiddensize}{n_h}
\newcommand{\nodenum}{|\nodeset|}
\newcommand{\nodei}{v_i}
\newcommand{\edgeij}{e_{i,j}}
\newcommand{\adjmat}{\textbf{A}}
\newcommand{\feati}{\featset_i}
\newcommand{\structspace}{\mathrm{S}}
\newcommand{\paramspace}{\mathrm{P}}
\newcommand{\substructspace}{\mathcal{S}}
\newcommand{\subparamspace}{\mathcal{P}}
\newcommand{\scnum}{m}
\newcommand{\pcnum}{n}
\newcommand{\neighbor}{\mathcal{N}}
\newcommand{\weight}{\textbf{W}}
\newcommand{\wij}{w_{i,j}}
\newcommand{\ns}{N_s}
\newcommand{\np}{N_p}
\newcommand{\loss}{\mathcal{L}}
\newcommand{\fij}{f_{i,j}}

\IEEEPARstart{N}{etwork} data and systems are ubiquitous \cite{zhang2018network,cui2018survey} in the real world including social network, document network, and biological network, \textit{etc}. Relationship modeling is of paramount importance for many network or graph data mining tasks (e.g., link prediction), which naturally desire flexible learning mechanisms to capture the discriminative pairwise node relationships at different levels, \textit{i.e.}, first-order and second-order neighborhoods. Recently, Graph Neural Networks (GNN) \cite{wu2020comprehensive,kipf2017semi} are developed to directly learn on networks or graphs, where nodes are allowed to incorporate high-order neighborhood relationships to generate node embeddings through the design of multiple graph convolution layers. For example, with a two-layer Graph Convolutional Networks (GCN) \cite{kipf2017semi}, the first and second GCN layers are able to respectively preserve the first-order and second-order neighborhood relationships in the embedding space. Due to the encouraging learning ability for graph-structured data in many domains, GNN has recently seen a plethora of successful real-world applications such as image recognition \cite{chen2019multi}, new drug discovery \cite{sun2019graph}, and traffic prediction \cite{zhao2019t}, \textit{ect}.

As of today, many GNN structures with diverse learning mechanisms have been proposed for node relationship modeling \cite{zhang2020deep,kriege2020survey}. For examples, GCN \cite{kipf2017semi} adopts a spectral-based convolution filter by which each node aggregates features from all direct neighborhoods. Graph Attention Network (GAT) makes each node aggregate features from all nodes on the network while learning to assign respective importance weights for different nodes. GraphSAGE \cite{hamilton2017inductive} learns a set of aggregation functions for each node to flexibly aggregate information from neighborhoods of different hops. Yet, developing a tailored learning architecture consisting of multiple GNN layers for a specific scenario (e.g., biological and physical network data) remains to be tricky, even for the neural network experts, because of two main reasons. First, each of the multiple GNN layers may prefer a different aggregation function (a.k.a. aggregator) to better capture neighborhood relationships in the respective order, \textit{i.e.}, GCN for the first-order while GAT for the second-order neighborhood relationships. Second, each specific aggregator alone may involve a number of structure selections such as activation function and the number of attention heads for GAT \cite{hamilton2017inductive}. As a result, to identify a superior model from the huge number of combinations of various components (e.g., aggregators and activation functions), one usually must apply tedious and laborious efforts for GNN structure tuning and optimization.

To automate the model selection process, Neural Architecture Search (NAS) is widespread used \cite{zoph2016neural,elsken2018neural} and has been a focal point of deep learning research in recent years. It seeks to find an optimal combination of architecture components from a well-defined searching space, where the resulting assembled neural model is suitable for a target problem. To date, massive efforts has been applied to searching the Convolutional Neural Network (CNN) architectures, which pushed forward the state-of-the-art in a number of significant benchmark tasks, \textit{e.g.}, image classification on CIFAR10/100 and ImageNet \cite{dong2019one,chen2019renas}. In contrast, very litter work has been done about GNN learning on graph-structured data. Two recent relevant works \cite{gao2019graphnas,zhou2019autognn} mainly focus on the reinforcement learning-based NAS and adopt a Recurrent Neural Network (RNN) as the controller to generate variable-length strings that describe the GNN structures. Despite the promising results, existing works are consistently challenged by the following two shortcomings:
\begin{itemize}
    \item \textbf{Invariable Hyper Parameters.} In addition to structures of GNN, a slight variation of hyper parameters (e.g., learning rate) could drastically impact the performance of a model with the converged structure. Existing methods optimizing only the structural variables with fixed hyper-parameter settings may end up with a suboptimal model.
    \item \textbf{High Computation and Low Scalability.} Training the recurrent network adds burden to the searching process. Both the training of controller and individual GNN model would demand extensive run-time computation resource. Furthermore, the controller typically generates candidate GNN structures and evaluates them in a sequential manner, which is difficult to scale to a large searching space. 
\end{itemize} 

\begin{figure}\label{fig1}
  \centering
  \includegraphics[width=0.45\textwidth]{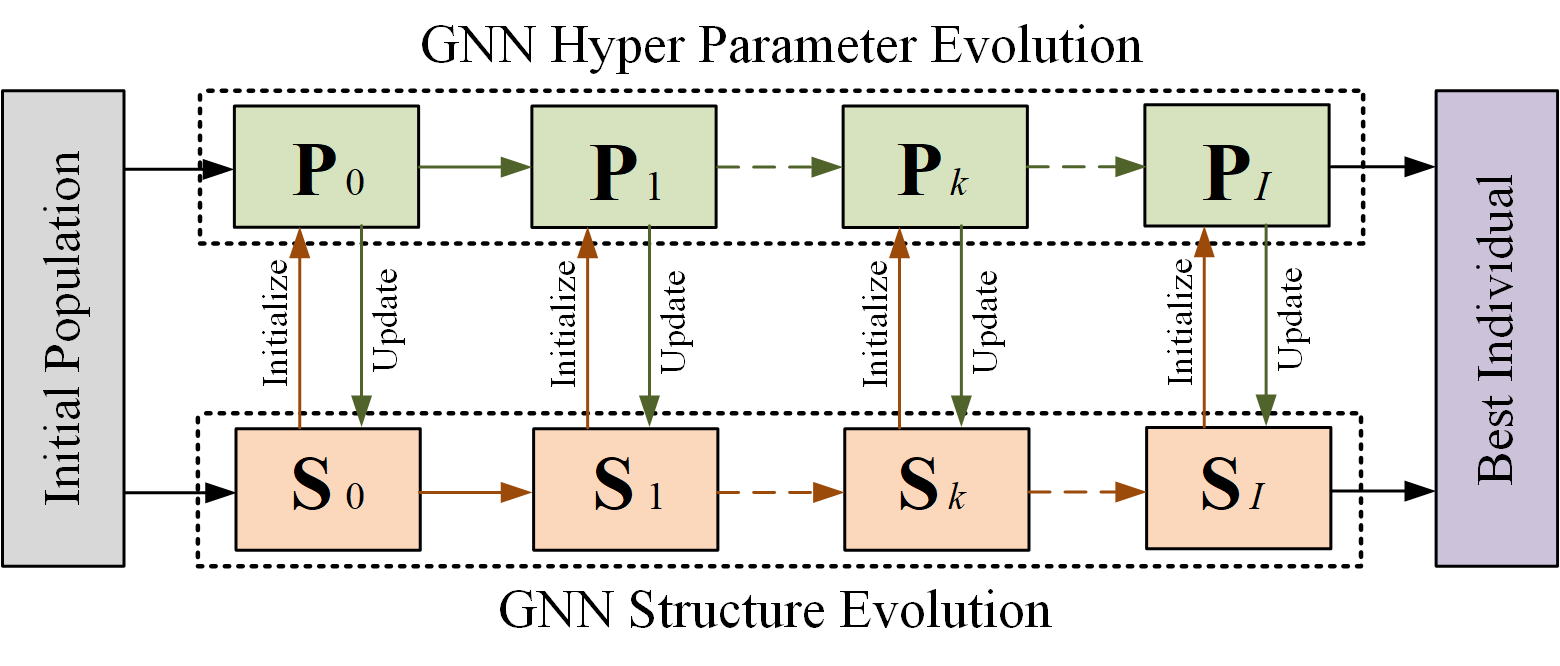}
  \vspace{-3mm}
  \caption{The proposed \text{\nickname} model for GNN architecture search. First, the population of GNN structures is initialized ($\structset_0$), where each individual is a multi-layer GNN with each layer constructed from randomly sampled components, \textit{i.e.}, aggregator, activation function, and hidden embedding size. Then, the population ($\paramset_0$) of GNN parameters \textit{w.r.t} $\structset_0$ is initialized and evolved to identify the optimal parameter setting (e.g., learning rate and dropout rate). Subsequently, the architecture evolution (from $\structset_0$ to $\structset_1$) is performed to optimize the GNN structures with the best parameter setting selected from $\paramset_0$. After $\numiter$ rounds of alternating evolution between the structure and parameter, it finally generates a GNN architecture with optimal structure and hyper parameter settings generated from $\structset_\numiter$ and $\paramset_\numiter$, respectively.}%
  \label{fig:motivation}
  \vspace{-5mm}
\end{figure}

To address aforementioned problems, this paper proposes a novel NAS framework termed \text{\nickname} through the evolution of individual models in a large GNN architecture space/population considering both model structures and learning parameters. However, jointly optimizing GNN structure and parameter is non-trivial, since the structure and parameter are dependent on each other in that a moderate change of hyper parameters could completely deteriorate the already fine-tuned model structures and vice verse. In the proposed model shown in Fig. \ref{fig:motivation}, we adopt an alternating evolution process to dynamically optimize both structures and parameters. In the structure evolution, each individual in the initial population represents a multi-layer GNN with randomly sampled components/parts for each layer. At each state $\structset_k$, to determine the optimal model parameters, we hold the population structures and meanwhile evolve to find the optimal parameters fitting the entire population well. Then, to find the optimal GNN structure upon a parameter setting, we hold the parameters and meanwhile evolve the entire population to optimize structures. 

Since both GNN structure and parameter can be evolved to fit each other dynamically, we expect to achieve a GNN architecture with both optimal structure and parameter settings for the target graph learning task (e.g., node classification). To the best of our knowledge, this is the first work to introduce and evaluate evolutionary architecture search for GNN models on graph-structured data. Extensive experiments and comparisons demonstrate that \text{\nickname} is capable of matching the state-of-the-art for both transductive and inductive node representation learning and classification. In addition, compared with existing reinforcement learning-based methods, our model can be easily scaled to large searching space since all individual models in each population are independent and thereby can be evaluated simultaneously.

In summary, the contribution of this work is as follows:
\begin{enumerate}
    \item We formulated a graph neural network architecture search problem under the evolutionary searching framework that seeks to optimize both model structures and hyper parameters.
    \item We proposed a novel evolutionary NAS framework called \text{\nickname} to automatically identify the optimal GNN architecture from a well-defined searching space. An alternating evolution process is performed to dynamically optimize GNN structures and parameters to fit each other. 
    \item We designed extensive experiments to demonstrate the effectiveness of evolutionary framework for GNN architecture search on both transductive and inductive graph representation learning and node classification. The results can provide guidance for other practitioners to select suitable graph neural models for a specific scenario.
\end{enumerate}

The rest of the paper is organized as follows. Section~\ref{sec:related} outlines related work about GNN and NAS. Section~\ref{sec:problem} defines the GNN architecture search problem. Section~\ref{sec:methods} establishes the underlying principles of the proposed \text{\nickname} model. Section~\ref{sec:experiment} evaluates the evolutionary GNN architecture search on both transductive and inductive graph learning tasks, and presents the comparative results on the benchmark datasets against baseline models. Finally, Section~\ref{sec:conclusion} concludes the paper while laying out possible directions of future work.

\section{Related Work} \label{sec:related}
This section first surveys current research on graph neural networks and the general neural architecture search, and then summarizes existing research targeted at the graph neural network architecture search and highlight their differences with our work in this paper.

\vspace{-3mm}
\subsection{Graph Neural Networks} 
Many real-world systems take the form of graph or network, \textit{i.e.}, social networks, citation networks, and biology molecular networks \cite{su2020network}. Different from grid-like data such as texts and images that are regular and sequential, networked data are irregular in that network nodes may have different numbers of unordered neighborhoods \cite{wu2020comprehensive}, causing the  failure of existing neural models such as CNN and RNN in the graph domain \cite{micheli2009neural,defferrard2016convolutional}. Recently, graph neural networks (GNN) as a family of neural network models were proposed to directly learn on graph-structured data \cite{kipf2017semi,niepert2016learning}. The main idea of GNN is to capture node relationships and features with carefully designed graph convolution kernels or filters \cite{defferrard2016convolutional}, where nodes are allowed to aggregate features from their respective neighborhood (e.g., first-order and second-order relationships) nodes iteratively. 

Naturally, flexible graph convolution kernels or feature aggregators are desired to efficiently model the complex node relationships in various graph systems and learning tasks, \textit{i.e.}, transductive and inductive network representation learning \cite{hamilton2017inductive, zhang2016tline}. To date, a significant number of graph neural kernel designs have been proposed \cite{kriege2020survey}. Gated graph neural network \cite{li2015gated} adopts a gated recurrent unit for neighborhood relationship modeling, where the hidden state of each node is updated by its previous hidden states and its neighboring hidden states. Chebyshev Spectral CNN (ChebNet) adapts the traditional CNN to learn on graphs by using the Chebyshev polynomial basis to represent the spectral filters \cite{defferrard2016convolutional}. GCN \cite{kipf2017semi} simplifies ChebNet architecture by using filters operating on 1-hop neighborhoods of the graph, where nodes in each GCN layer only aggregate features from their direct neighbors. Diffusion CNN (DCNN) \cite{atwood2016diffusion} regards graph convolutions as a diffusion process. It assumes information is transferred from one node to each of its neighborhoods following a transition probability distribution. In addition to convolution filters that treat neighborhood nodes as equally important, many works demonstrate that attention-based filters could be useful. For example, Graph Attention Networks (GAN) \cite{velivckovic2017graph} introduce an attention mechanism to determine the importance of neighborhoods to the center node during feature aggregation.

Although diverse graph convolution filters and feature aggregators have been proposed to achieve new-record performance in many real-world applications (e.g., node classification and link prediction), it is prohibitive to identify a GNN model which is suitable for all kinds of networked data and systems \cite{gao2019graphnas,zhou2019autognn}. In general, nodes build relationships with each other in different granularity, \textit{i.e.}, direct and indirect neighborhood relations, which intuitively demands varying graph filters for different feature aggregations and relationship modeling. For example, GraphSAGE \cite{hamilton2017inductive} tries to train a set of aggregator functions that learn to aggregate feature information from a node’s local neighborhood, where each aggregator function aggregates information from a different number of hops, or search depth, away from a given node. 

\vspace{-3mm}
\subsection{General Neural Architecture Search} 
Neural Architecture Search (NAS) is a fundamental step in automating the machine learning process, which has been successfully applied in many real-world applications such as image segmentation \cite{liu2019auto} and text processing \cite{wang2020textnas}. NAS aims to design a model architecture with the best performance using limited computing resources in an automated way with litter or no human intervention. Most of existing works can be roughly classified into three categories, including reinforcement learning, Bayesian, and evolutionary optimizations \cite{elsken2018neural,wistuba2019survey}.

Reinforcement Learning (RL), functioning as a model architecture selection controller, has been extensively used in automating CNN model designs \cite{jaafra2019reinforcement}. Zoph et al. \cite{zoph2016neural} first used a RNN to generate the string description of CNN model and then trained this RNN with RL to maximize the expected accuracy (e.g., image recognition) of the generated model. Baker et al. \cite{baker2016designing} proposed a RL-based meta-modeling algorithm called MetaQNN which incorporates a novel Q-learning agent whose goal is to discover CNN architectures that perform well on a given machine learning task with no human intervention. Above methods often design and train each network from scratch during the exploration of the architecture space. To enable more efficient training, Cai et al. \cite{cai2018efficient} proposed a RL-based method where weights for historical network models can be reused to train the current model. However, a noticeable limitation for RL-based NAS is the low scalability, especially when the searching space is very large, since the candidate models are sequentially dependent on each other for progressive optimization. 

Bayesian Optimization (BO) is a family of algorithms that build a probability model of the objective function determining the best expected neural network architecture \cite{he2019automl}. Early work proposed using the tree-based frameworks such as random forest and tree Parzen estimators \cite{bergstra2011algorithms}. For example, Bergstra et al. proposed a non-standard Bayesian-based optimization algorithm TBE, which uses tree-structured Parzen estimators to model the error distribution in a non-parametric way. Hutter et al. \cite{hutter2011sequential} proposed SMAC which is a tree-based algorithm that uses random forests to estimate the error density. Gaussian processes are also widely used in the Bayesian optimization. Kandasamy et al. \cite{kandasamy2018neural} proposed a Gaussian process based BO framework for searching multi-layer perceptron and convolutional neural network architectures, which is performed sequentially where at each time step all past model evaluation results are viewed as posterior to construct an acquisition function evaluating the current model.

Evolutionary Algorithm (EA) performs an iterative genetic population-based meta-heuristic optimization process, which is a mature global optimization method with high robustness and wide applicability \cite{elsken2018neural,he2019automl}. EA-based NAS has been widely used for identifying suitable CNN model for a specific task such as image denoising, in-painting and super-resolution \cite{ho2020neural}. Different EA-based algorithms may use different types of genome encoding methods for the neural model architectures. For example, Genetic-CNN \cite{xie2017genetic} proposed to represent each network structure in a fixed-length binary string, where each element in the string corresponds to a specific kind of operation. Masanori et al. \cite{suganuma2017genetic} proposed to use Cartesian genetic programming to represent the CNN structure and connectivity, which can represent variable-length network structures and skip connections.

\vspace{-3mm}
\subsection{Graph Neural Network Architecture Search} 
Most existing work focuses on NAS of CNN models learning on grid-like data such as texts and images. For NAS of GNN models evaluating on graph structured data, very litter work has been done so far. GraphNAS \cite{gao2019graphnas} proposed a graph neural architecture search method based on the reinforcement learning. It first uses a recurrent network to generate variable-length strings to describe GNN architectures, and then trains the recurrent network with reinforcement learning to maximize the expected accuracy of the generated architectures on a validation data set. Auto-GNN \cite{zhou2019autognn} follows a similar architecture searching paradigm as GraphNAS while proposing a parameter sharing strategy that enables homogeneous architectures to share parameters. These works all focus on the GNN structures (e.g., aggregators and activation functions) optimization with fixed learning parameters. However, GNN structures and hyper parameters would impact each other in that the moderate change of learning parameters (e.g., learning rate) could severely degrade the accuracy of the optimal GNN architecture achieved by existing methods.

In comparison, studying from a different line we aim to evaluate the effectiveness of evolutionary method for GNN architecture search, and propose a novel framework through the evolution of individual models in a large GNN architecture space. In addition, other than optimizing the GNN structures as existing methods, we propose to evolve and optimize both GNN structures and learning parameters given that they may impact each other in the searching process. More specific, we propose a two-phase encoding scheme which uses two strings to respectively represent the GNN structures and hyper parameters. By this way, we are able to evolve and optimize both structures and learning parameters to fit each other.

\section{Problem Definition} \label{sec:problem}
In this paper, the objective is to identify the optimal GNN architecture by NAS for network representation learning (a.k.a network embedding) by doing a semi-supervised node classification training. Formally, the tasks of network embedding and graph NAS are defined as follows.

\vspace{1mm}
\noindent\textbf{Definition 1 (Network Embedding).} The target network can be represented by $\graph=(\nodeset,\edgeset,\featset)$, where $\nodeset=\{\nodei\}_{i=1,\cdots,\nodenum}$ is a set of $\nodenum$ unique nodes, $\edgeset=\{ \edgeij \}_{ {i,j}=1,\cdots,\nodenum;~i \neq j }$ is a set of edges that can be represented by a $\nodenum \times \nodenum$ adjacency matrix $\adjmat$, with $\adjmat_{i,j}=1$ if $\edgeij \in \edgeset$, or $\adjmat_{i,j}=0$ otherwise. $\featset$ is a matrix $\spacer^{\nodenum \times \featnum}$ containing all $\nodenum$ nodes with their associated features, \textit{i.e.}, $\feati \in \spacer^{\featnum}$ is the feature vector of node $\nodei$, where $\featnum$ is the number of unique node features. The task of network embedding is to learn a mapping $f:\graph \rightarrow\{\embedh_i\}_{i=1,\cdots,\nodenum}$ by preserving network topology and node features, where $\embedh_i \in \spacer^{\embedsize}$ represents the low-dimensional vector representation of node $\nodei$, and $\embedsize$ is the embedding vector's dimension. $f$ can be the GNN model identified by NAS in this paper. The mapping is learned in a semi-supervised manner, \textit{i.e.}, labels for a small part of nodes are known, where the node embedding vectors are trained to predict their respective labels. 

\vspace{1mm}
\noindent\textbf{Definition 2 (Graph Neural Architecture Search).} For the graph NAS task in this paper, the GNN structure space $\structspace\in \spacer^{|\substructspace_1| \times |\substructspace_2|\times \cdots \times |\substructspace_\scnum|}$ and GNN learning parameter space $\paramspace \in \spacer^{|\subparamspace_1| \times |\subparamspace_2|\times \cdots \times |\subparamspace_\pcnum|}$ have been given, where $\substructspace_{i=1,2,\cdots,\scnum}\in \spacer^{|\substructspace_i|}$ is the set of candidate choices for the $i^{th}$ structure component (e.g., GNN aggregator) and $\subparamspace_{j=1,2,\cdots,\pcnum}\in \spacer^{|\subparamspace_j|}$ is the set of candidate choices for the $j^{th}$ learning parameter. $\scnum$ and $\pcnum$ are respectively the numbers of structure components and learning parameters required to build a GNN model. The task of graph NAS is to identify the optimal choices or value specifications (e.g., $\substructspace^{best}_i$ and $\subparamspace^{best}_j$) for each structure component $\substructspace_i$ and learning parameter $\subparamspace_j$, such that the constructed GNN model $f=\left\{\substructspace^{best}_1,\substructspace^{best}_2,\cdots,\substructspace^{best}_\scnum;\subparamspace^{best}_1,\subparamspace^{best}_2,\cdots,\subparamspace^{best}_\pcnum\right\}$ could achieve the optimal embedding performance learning on the target network $\graph$.

\begin{figure}
  \centering
    \includegraphics[width=0.42\textwidth]{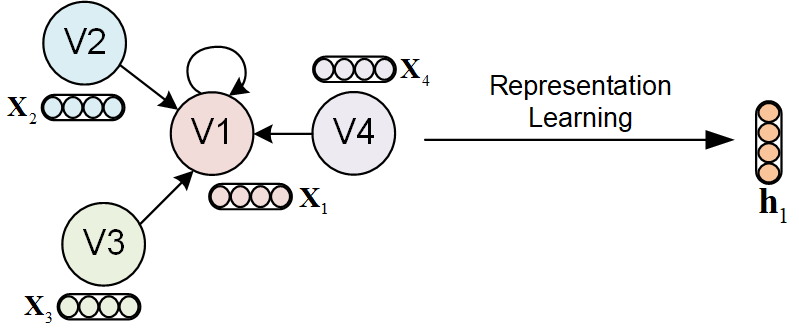}
  \caption{The representation learning scheme of GNN models.}%
  \label{fig:fig2}
  \vspace{-3mm}
\end{figure}

\begin{figure*}[t]
  \centering
    \includegraphics[width=0.92\textwidth]{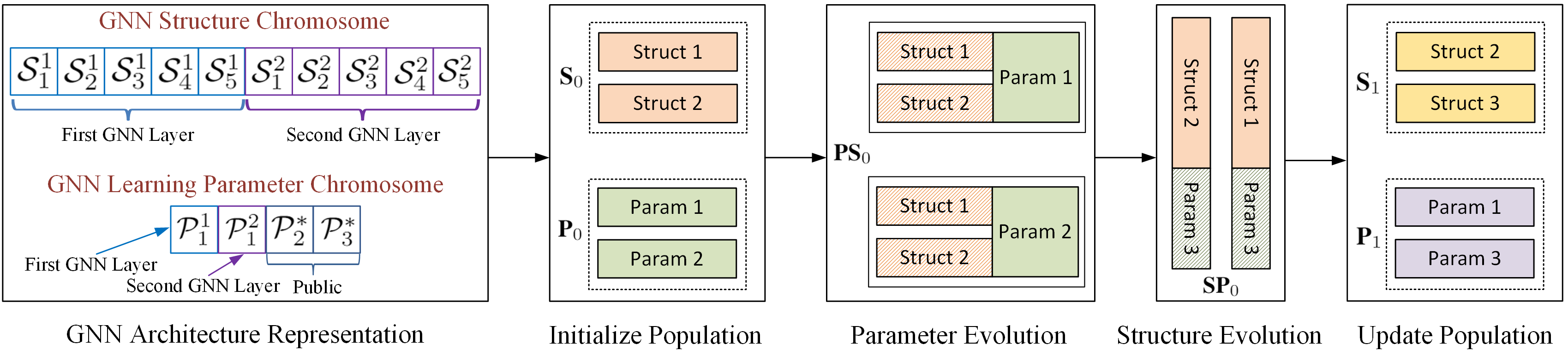}
  \caption{The proposed \text{\nickname} model for GNN architecture optimization. For simplicity, we demonstrate one evolution step of a two-layer GNN, where both structure and parameter populations have two individuals. \textbf{First}, the GNN structure population ($\structset_0$) and parameter population ($\paramset_0$) are randomly initialized, where each structure or parameter individual is represented with a respective string/chromosome. \textbf{Second}, an intermediate population $\paramset\structset_0$ is constructed to evolve parameters with structures fixed, \textit{i.e.}, assume a better parameter individual \textit{Param 3} is produced to replace \textit{Param 2}. \textbf{Then}, another intermediate population $\structset\paramset_0$ is constructed to evolve structures with parameters fixed, \textit{i.e.}, assume a better structure individual \textit{Struct 3} is produced to replace \textit{Struct 1}. \textbf{Finally}, both structure and parameter populations are updated.}%
  \label{fig3}
\end{figure*}

\section{Preliminaries}
To support the proposed graph NAS framework, this section briefly introduces preliminary knowledge about graph neural networks and genetic algorithm.

\subsection{Graph Neural Networks}
GNN is a family of neural network models that can directly incorporate graph topology and node features for efficient low-dimensional node representation learning. As shown in Fig. 2, the main idea for GNN models is that each node $\nodei$ generates the representation $\embedh_i$ by aggregating features from its neighborhoods (e.g., the first-order neighbors in this paper). Typically, the following five GNN structure components are involved in above representation learning scheme: 
\begin{itemize}
    \item[1.] \textbf{Attention Function ($\substructspace_1$).} While each node aggregates features from its neighbors, different neighborhood nodes may have different contributions aligned with the affinities between nodes \cite{velivckovic2017graph}. The attention function aims to learn an importance weight $w_{ij}$ for each edge relationship $\edgeij$ between the two corresponding nodes $\nodei$ and $v_j$. $\substructspace_1$ denotes the set of candidate attention functions.
    \item[2.] \textbf{Attention Head ($\substructspace_2$).} Instead of applying the attention function once, studies \cite{vaswani2017attention} show that it is beneficial to perform multiple attentions independently. Multi-head attention allows the model to jointly attend to features from different node representation subspaces. The multiple representation outputs by multi-head attention for each node $\nodei$ are then concatenated or averaged to generate the final representation $\embedh_i$. $\substructspace_2$ denotes the set of candidate attention head numbers.
    \item[3.] \textbf{Aggregation Function ($\substructspace_3$).} Each node $\nodei$ may have multiple neighborhood nodes, thus an aggregation function (e.g., averaging operation) is required to combine features from multiple neighbors to form the representation $\embedh_i$. $\substructspace_3$ denotes the set of candidate aggregation functions.
    \item[4.] \textbf{Activation Function ($\substructspace_4$).} After deriving the representation $\embedh_i$ for node $\nodei$, a non-linear activation function (e.g., ReLu and Sigmoid) is usually applied to smooth $\embedh_i$ or transform $\embedh_i$ as probability vector for node classification. $\substructspace_4$ denotes the set of candidate activation functions.
    \item[5.] \textbf{Hidden Unit ($\substructspace_5$).} The hidden unit controls the dimension of the representation $\embedh_i$ for node $\nodei$. $\substructspace_5$ denotes the set of candidate dimension choices.
\end{itemize}

Based on above definitions, the first-layer learning structure of a GNN model (e.g., assume it is instantiated and indexed by 1) can be represented as an action string by $\structspace_1=\left\{\substructspace_1^1, \substructspace_2^1, \substructspace_3^1, \substructspace_4^1, \substructspace_5^1\right\}$. Formally, the representation $\embedh_i\in\spacer^{\substructspace_5^1}$ for node $\nodei$ can be computed as:
\begin{equation}
    \embedh_i = \mathop{\parallel}^{\substructspace_2^1}_{l=1}\substructspace_4^1\left(\mathop{\substructspace_3^1}_{j\in\neighbor_i}\left(\substructspace_1^1(\featset_i, \featset_j, \weight^l)\weight^l\featset_j\right)\right)
\end{equation}
where $\parallel$ represents concatenation, $\neighbor_i$ represents the set of direct (e.g., first-order) neighborhoods of node $\nodei$, and $\weight^l\in \spacer^{\featnum}$ is the learnable weight matrix for the $l^{th}$ attention head. One can stack multiple GNN layers to form a multi-layer GNN model. At the $k^{th}$ layer, assume the action string is instantiated as $\structspace_k=\left\{\substructspace_1^k, \substructspace_2^k, \substructspace_3^k, \substructspace_4^k, \substructspace_5^k\right\}$, then the output representation $\embedh_i^{(k)}\in \spacer^{\substructspace_5^k}$ is written as: 
\begin{equation}
    \embedh_i^{(k)} = \mathop{\parallel}^{\substructspace_2^k}_{l=1}\substructspace_4^k\left(\mathop{\substructspace_3^k}_{j\in\neighbor_i}\left(\substructspace_1^k(\embedh_i^{(k-1)}, \embedh_j^{(k-1)}, \weight^l)\weight^l\embedh_j^{(k-1)}\right)\right)
\end{equation}
where $\embedh_i^{(k-1)}$ is the output representation by $(k-1)^{th}$ GNN layer and $\weight^l\in \spacer^{\substructspace_5^{(k-1)}}$ represents the corresponding learnable weight matrix. If $k$ indicates the last GNN layer, the aggregation function $\substructspace_3^k$ will be the \textit{averaging} operation, meaning averaging the representations generated by all $\substructspace_2^k$ attention heads. Then, the final representation for node $\nodei$ is written as:
\begin{equation}
    \embedh_i^{(k)} = \substructspace_4^k\left(\frac{1}{\substructspace_2^k}\sum_{j\in\neighbor_i}\left(\substructspace_1^k(\embedh_i^{(k-1)}, \embedh_j^{(k-1)}, \weight^l)\weight^l\embedh_j^{(k-1)}\right)\right)
\end{equation}

The weight matrix $\weight_l$ at each GNN layer is trained in a semi-supervised manner, \textit{i.e.}, by performing node classification \cite{kipf2017semi,velivckovic2017graph} optimized with the gradient decent algorithm. 

\subsection{Genetic Algorithm}
Genetic Algorithm (GA) is a kind of evolutionary algorithm motivated by the principle of natural selection and genetics \cite{sheikh2008genetic}. The search space is a major ingredient for all GAs, which is encoded in the form of strings known as \textit{chromosomes} or \textit{individuals}, and a collection of such strings form a \textit{population}. A \textit{chromosome} is composed of a sequence of elements called \textit{genes}, which encode the solution of a target problem. Initially, a random population is created representing different individual solutions for the target problem. A \textit{fitness} value is associated with each individual to indicate its goodness in the population. GA optimizes the population and tries to find the global optimal solution through a standard evolution procedure as follows:
\begin{enumerate}
	 \item[1.] \textbf{Initialize}(population)
	 \item[2.] \textbf{Evaluate}(population)
	 \item[3.] \textbf{While}(stopping condition not satisfied):
      \begin{enumerate} 
         \item \textbf{Selection}(population)
         \item \textbf{Crossover}(population)
         \item \textbf{Mutate}(population)
         \item \textbf{Evaluate}(population)
         \item \textbf{Update}(population)
      \end{enumerate}
      \item[4.] \textbf{Return} the best individual in the population
\end{enumerate}
where the evaluation step aims to calculate the fitness of each individual, the selection step aims to choose some individuals from the entire population as parents for mating, the crossover step describes how parental individuals switch information (e.g., swap the genes) and produce next generations (e.g., new individuals), the mutation step aims to introduce diversity in the population by randomly altering a gene from new individuals conditioned on the mutation probability, and finally it update the population by adding the new individuals. 

\section{The Proposed Method} \label{sec:methods}
This section presents a Genetic Graph Neural Network (Genetic-GNN) NAS framework to evolve the GNN architecture. As shown in Fig. 3, Genetic-GNN can be organized in three main components set to optimize both the GNN structure and learning parameters, including GNN architecture representation \& population initialization, GNN learning parameter evolution, and GNN structure evolution. We elaborate the three components in the following.

\subsection{Architecture Representation \& Population Initialization}
As discussed in previous sections, the optimizations of GNN structure and learning parameters are dependent on each other. Existing works optimize only GNN structures may end up with a suboptimal searched model since the change of learning parameters could severely degrade the fine-tuned GNN structure. Therefore, we recommend evolving both GNN structure and learning parameters for reliable NAS.

In this paper, we are specifically interested in optimizing the following three types of learning parameter, although Genetic-GNN is a general framework which is flexible to include other significant parameters:
\begin{itemize}
    \item[1.] \textbf{Dropout Rate ($\subparamspace_1$).} Overfitting is a common issue when training neural network models. Dropout is a technique for addressing this problem, which meanwhile helps to reduce the training complexity for large networks \cite{srivastava2014dropout}. The key idea is to randomly drop units (along with their connections) from the neural network during training. $\subparamspace_1$ denotes the set of candidate dropout rate values.
    \item[2.] \textbf{Weight Decay Rate ($\subparamspace_2^*$).} Similar to the dropout, weight decay (e.g., $L_2$ norm regularization) is a widely used technique to decrease the complexity and meanwhile increase the generalization ability of neural network models by limiting the growth of model weights. $\subparamspace_2^*$ denotes the set of candidate weigh decay rate values.
    \item[3.] \textbf{Learning Rate ($\subparamspace_3^*$).} Learning rate determines how fast the loss changes every time while training a neural model based on the gradient decent algorithm. A larger learning rate could cause the model to converge too quickly to a suboptimal solution, whereas a smaller learning rate could cause the optimization to converge too slowly. $\subparamspace_3^*$ denotes the set of candidate learning rate values.
\end{itemize}
The weigh decay is usually applied to GNN model weights at the first layer and the learning rate is set for training the entire GNN model. Therefore, the learning parameters $\subparamspace_2^*$ and $\subparamspace_3^*$ are set once and maintain public through multiple layers of a GNN model.

Assume we search the optimal architecture for a two-layer GNN model, as shown in Fig. 3, the GNN structure can be represented by a string/chromosome:
\begin{equation}
    \left\{\substructspace_1^1, \substructspace_2^1, \substructspace_3^1, \substructspace_4^1, \substructspace_5^1, \substructspace_1^2, \substructspace_2^2, \substructspace_3^2, \substructspace_4^2, \substructspace_5^2\right\}
\end{equation}
where the first and second GNN layers are indexed by 1 and 2, respectively. Similarly, the GNN learning parameters can be represented by a string/chromosome:
\begin{equation}
    \left\{\subparamspace_1^1, \subparamspace_1^2, \subparamspace_2^*, \subparamspace_3^*\right\}
\end{equation}
where public parameters in the two GNN layers are indexed by the notation *. While evaluating the network embedding performance (e.g., fitness of individuals), the structure and parameter strings need to be combined to form the entire GNN architecture (e.g., the mapping function $f$):
\begin{equation}
   f = \left\{\substructspace_1^1, \substructspace_2^1, \substructspace_3^1, \substructspace_4^1, \substructspace_5^1, \substructspace_1^2, \substructspace_2^2, \substructspace_3^2, \substructspace_4^2, \substructspace_5^2; \subparamspace_1^1, \subparamspace_1^2, \subparamspace_2^*, \subparamspace_3^*\right\}
\end{equation}

For a given graph $\graph$, two datasets are created, including a training node set $D_{train}$ and a validation node set $D_{val}$. The candidate GNN model $f$ is trained on $D_{train}$ by minimizing the semi-supervised node classification loss:
\begin{equation}
    \loss_f=-\sum\nolimits_{\nodei\in D_{train}}\mathbf{Y}_i\ln\embedh_i
\end{equation}
where $\mathbf{Y}_i$ is the one-hot label indicator vector for node $\nodei$. Then, the classification accuracy of $f$ is computed on $D_{val}$.

\begin{table}
\renewcommand{\arraystretch}{1.60}
\centering
\caption{The Search Space for Structure Components.}
\label{tab:data}
    \begin{tabular}{c|c}
    \hline
    \textbf{Component} & \textbf{Search Space}  \\
    \hline
    \hline
    $\substructspace_1$ & listed in Table II\\
    \hline
    $\substructspace_2$ & 1, 24, 6, 8, 16\\
    \hline
    $\substructspace_3$ & ``sum", ``mean-pooling", ``max-pooling", ``mlp"\\
    \hline
    $\substructspace_4$ & \makecell{``sigmoid", ``tanh", ``relu", ``linear", ``softplus",\\``leaky\_relu", ``relu6", ``elu"}\\
    \hline
    $\substructspace_5$ & 4, 8, 16, 32, 64, 128, 256\\
    \hline
    \end{tabular}
\end{table}

\begin{table}
\renewcommand{\arraystretch}{1.60}
\centering
\caption{The Search Space of Structure Component $\substructspace_1$.}
\label{tab:data}
    \begin{tabular}{c|c}
    \hline
    \textbf{Search Space} & \textbf{Definition}  \\
    \hline
    \hline
    const & $\wij=1$\\
    \hline
    gcn & $\wij=\frac{1}{\sqrt{\neighbor_i\neighbor_j}}$\\
    \hline
    gat & $\wij=leaky\_relu(\weight^l*\embedh_i+\weight^l*\embedh_j)$\\
    \hline
    sym-gat & $\wij=\wij+w_{j,i}$ based on gat\\
    \hline
    cos & $\wij=cos(\weight^l*\embedh_i,\weight^l*\embedh_j)$\\
    \hline
    linear & $\wij=tanh(sum(\weight^l*\embedh_j))$\\
    \hline
    gene-linear & \makecell{$\wij=\weight^a*tanh(\weight^l*\embedh_i+\weight^l*\embedh_j)$, \\ where $\weight^a$ is a trainable weight matrix} \\
    \hline
    \end{tabular}
\end{table}

\begin{table}
\renewcommand{\arraystretch}{1.60}
\centering
\caption{The Search Space for Learning Parameters.}
\label{tab:data}
    \begin{tabular}{c|c}
    \hline
    \textbf{Parameter} & \textbf{Search Space}  \\
    \hline
    \hline
    $\subparamspace_1$ & 0.05, 0.1, 0.2, 0.3, 0.4, 0.5, 0.6\\
    \hline
    $\subparamspace_2^*$ & 5e-4, 8e-4, 1e-3, 4e-3\\
    \hline
    $\subparamspace_3^*$ & 5e-4, 1e-3, 5e-3, 1e-2\\
    \hline
    \end{tabular}
\end{table}

Following the literature \cite{gao2019graphnas}, the candidate choices or search space for each GNN structure component are summarized in Table I and Table II. Similarly, we define the search space for each learning parameter which is summarized in Table III. Based on the chromosomes (e.g., Eq. (4) and Eq. (5)) and their respective search spaces (e.g., Table I and Table III), the GNN structure population $\structset_0=\left\{struct_i\right\}_{i=1,\cdots,\ns}$ and learning parameter population $\paramset_0=\left\{param_j\right\}_{j=1,\cdots,\np}$ are respectively created and initialized, \textit{i.e.}, feeding each structure component in Eq. (4) and learning parameter in Eq. (5) with values randomly selected from their respective search spaces, where $\ns$ and $\np$ are the population sizes (e.g., number of individuals). We use $\fij=\left\{struct_i;param_j\right\}$ to represent a candidate GNN model and its classification accuracy is $Acc(\fij)$, where $struct_i\in \structset_0$ and $param_j\in \paramset_0$. In the following sections, we adopt an alternating evolution procedure to evolve $\structset_0$ and $\paramset_0$, aiming to identify the optimal $f$ for learning a target graph.

\begin{figure}
  \centering
    \includegraphics[width=0.4\textwidth]{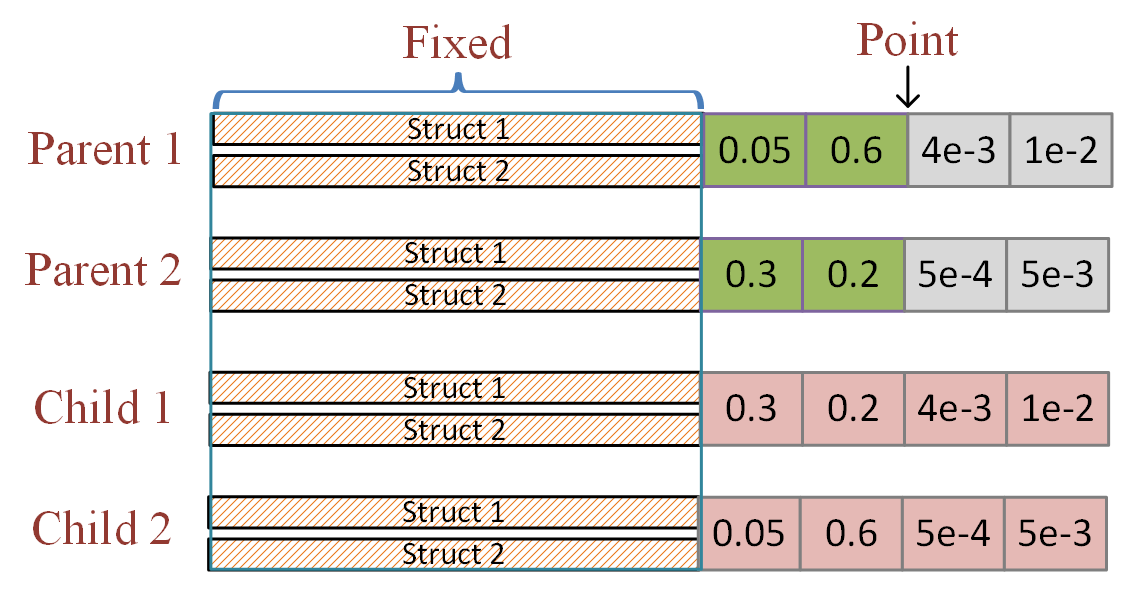}
  \caption{An example to show the crossover process between two parents, where the point index for learning parameter crossover is 2.}%
  \label{fig:fig4}
\end{figure}

\subsection{GNN Learning Parameter Evolution}
The parameter evolution component aims to evolve $\paramset_k$ to optimize and identify the optimal parameter setting for the corresponding GNN structure population $\structset_k$ before evolving to the next structure population $\structset_{k+1}$, where the goodness/fitness of a particular parameter setting should be measured regarding all individuals in the structure population. To this end, as shown in Fig. 3 (e.g., $k=0$), we combine $\structset_0$ and $\paramset_0$ to create an intermediate population $\paramset\structset_0=\left\{PS_j\right\}_{j=1,\cdots,\np}$, where each individual $PS_j=\left\{f_{i,j}\right\}_{i=1,\cdots,\ns}$ encapsulates a set of GNN models with the same learning parameter setting $param_j\in \paramset_0$, and different individuals have the same GNN structure settings in $\structset_0$. The fitness of $PS_j$ is computed as:
\begin{equation}
    fitness(PS_j)=\alpha Acc(f_{b,j})+(1-\alpha)\frac{1}{\ns}\sum_{i=1}^{\ns}Acc(\fij)
\end{equation}
where $f_{b,j}=\left\{struct_b;param_j\right\}\leftarrow\text{argmax}_{\fij\in PS_j}Acc(\fij)$ is the best individual with highest classification accuracy $Acc(f_{b,j})$ in the population. The last term of Eq. (8) calculates the average classification accuracy over all individuals. The motivation is that we seek to find a parameter setting $param_j$ that fits the entire structure population $\structset_0$ while simultaneously considering its fit to the best structure individual $struct_b\in \structset_0$. $\alpha$ is a balance parameter which allows us to adjust the importance between these two sides for flexible fitness calculation.

Then, based on the standard GA algorithm, we evolve $\paramset\structset_0$ to optimize the learning parameter $\paramset_0$ by holding the GNN structure $\structset_0$, which includes the selection, crossover, mutation, evaluation and updating steps. For example, Fig. 4 shows the crossover step between two selected parents, where the crossover only happens for the learning parameters with the GNN structures fixed. The parameter evolution (e.g., $\paramset_0$ evolves to $\paramset_{1}$) finally identifies and outputs the parameter individual with highest fitness calculated by Eq. (8), \textit{i.e.}, the $param_3$ shown in Fig. 3. 

\subsection{GNN Structure Evolution}
Once the optimal learning parameter $param_j\in \paramset_{k+1}$ for $\structset_k$ has been identified, the structure evolution component aims to evolve $\structset_k$ to identify the optimal GNN structure. For this purpose, as the case (e.g., $k=0$) shown in Fig. 3, an intermediate population $\structset\paramset_0=\left\{SP_i\right\}_{i=1,\cdots,\ns}$ is created by concatenating the optimal parameter individual with each structure individual $struct_j\in \paramset_0$, where $SP_i=f_{i,j}$ 
indicates an individual GNN model with its fitness calculated as:
\begin{equation}
    fitness(SP_i)=Acc(f_{i,j})
\end{equation}

Similarly, we evolve $\structset\paramset_0$ to optimize the GNN structure $\structset_0$ (e.g., $\structset_0$ evolves to $\structset_1$) by holding the learning parameter $\paramset_0$ based on the standard GA algorithm. For example, Fig. 5 shows the crossover process by altering only the structural parts of the two parents while keeping the learning parameters fixed. 

\begin{figure}
  \centering
    \includegraphics[width=0.44\textwidth]{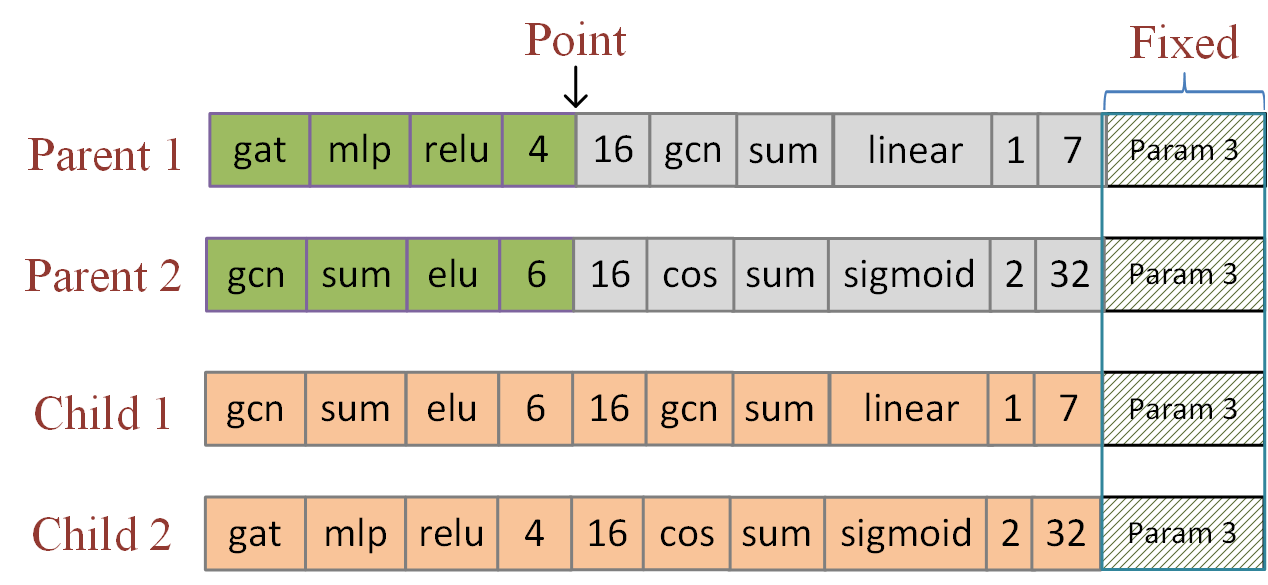}
  \caption{An example to show the crossover process between two parents, where the point index for structure crossover is 4.}%
  \label{fig:fig5}
\end{figure}

\subsection{Algorithm Explanation}
The learning parameter evolution and GNN structure evolution proceed in an alternating manner. Before evolving the structure population $\structset_k$, the parameter population $\paramset_k$ is evolved to identify the optimal learning parameters fitting $\structset_k$ and meanwhile $\paramset_k$ evolves to $\paramset_{k+1}$. Subsequently, the structure population $\structset_k$ is evolved to optimize the GNN structures and meanwhile $\structset_k$ evolves to $\structset_{k+1}$. Above alternating process is performed iteratively to finally achieve a multi-layer GNN architecture with both optimal structures and learning parameters. The training procedure of \text{\nickname} is summarized in Algorithm 1, where $K_s$ and $K_p$ are number of generations for structure evolution and parameter evolution, respectively. For the evolution of intermediate populations $\paramset\structset_k$ and $\structset\paramset_k$, since all individual GNN models are independent and can be evaluated simultaneously, \text{\nickname} is able to scale to large search space and individual population. 

\section{Experiment} \label{sec:experiment}
Following literature \cite{gao2019graphnas,zhou2019autognn}, we test the performance of \text{\nickname} on both transductive and inductive node representation learning by performing the supervised node classification training.

\subsection{Dataset}
The four datasets used in the two tasks are summarized in Table IV. Three benchmark citation networks including Cora, Citeseer, and Pubmed are used for transductive node representation learning, where 20 nodes per lass for training (e.g., the train set $D_{train}$), 500 nodes for validation (e.g., the validation $D_{val}$) and 1,000 nodes for testing. 

We use a protein-protein interaction (PPI) dataset for inductive node representation learning, which contains 20 graphs for training, 2 graphs for validation and 2 graphs for testing. Each graph have 2,372 nodes on average, and each node has 50 features including positional gene sets, motif gene sets and immunological signatures. Each node corresponds to multiple labels from the total of 121 classes.

\begin{algorithm}[t]
\caption{Training procedure of the \text{\nickname} model}
\label{alg:loop}
\begin{algorithmic}[1]
\Require{The target graph $\graph$, train set $D_{train}$ and validation set $D_{val}$}
\Ensure{The optimal GNN architecture and the learned embedding vector $\textbf{h}_{v_i}$ for each node $v_i \in \mathbf{V}$} 
\Statex
\State Initialize the structure population $\structset_0$
\State Initialize the learning parameter population $\paramset_0$
\Procedure{ArchitectEvolving}{$D_{train}$, $D_{val}$, $\structset_0$, $\paramset_0$, $K_s$, $K_p$}
    \For{$i \gets 1$ to $K_s$}
        \State{Construct intermediate GNN model population $\paramset\structset_i$}
        \For{$j \gets 1$ to $K_p$}
        \State{\textbf{Selection}($\paramset\structset_i$)}
        \State{\textbf{Crossover}($\paramset\structset_i$)}
        \State{\textbf{Mutate}($\paramset\structset_i$)}
        \State{\textbf{Evaluate}($\paramset\structset_i$)}
        \State{\textbf{Update}($\paramset\structset_i$)}
        \EndFor
        \State{Construct intermediate GNN model population $\structset\paramset_i$}
        \State{\textbf{Selection}($\structset\paramset_i$)}
        \State{\textbf{Crossover}($\structset\paramset_i$)}
        \State{\textbf{Mutate}($\structset\paramset_i$)}
        \State{\textbf{Evaluate}($\structset\paramset_i$)}
        \State{\textbf{Update}($\structset\paramset_i$)}
    \EndFor
\EndProcedure
\end{algorithmic}
\end{algorithm}

\begin{table}
\renewcommand{\arraystretch}{1.3}
\centering
\caption{Benchmark network datasets characteristics.}
\begin{small}
\label{tab:data}
    \begin{tabular}{c|c|c|c|c}
    \hline
    Items & Cora & Citeseer & PubMed & PPI \\
    \hline
    \# Nodes & $2,708$  & $3,327$ & $19,717$ &$56,944$ \\
    \hline
    \# Features & $1,433$ & $3,703$ & $500$ & $50$ \\
    \hline
    \# Classes & $7$ & $6$ & $3$ & $121$ \\
     \hline
    \# Training Nodes & $140$ & $120$ & $60$ & $44,906$ \\
     \hline
    \# Validation Nodes & $500$ & $500$ & $500$ & $6,514$ \\
     \hline
    \# Testing Nodes & $1,000$ & $1,000$ & $1,000$ & $5,524$ \\
    \hline
    \end{tabular}
\end{small}
\end{table}

\subsection{Baseline}
We compare \text{\nickname} with the following state-of-the-art methods which adopt either handcrafted GNN architecture or GNN architecture search.

\textit{Methods based on Handcrafted GNN Architecture}:
\begin{itemize}
    \item \textbf{Chebyshev} \cite{defferrard2016convolutional} adapts the traditional CNN to learn on graphs by using the Chebyshev polynomial basis to  represent the spectral CNN’s filters.
    \item \textbf{GCN} \cite{kipf2017semi} is a two-layer GCN architecture, where each node generates representation by adopting a spectral-based convolutional filter to recursively aggregate information from all its direct neighbors.
    \item \textbf{GraphSAGE} \cite{hamilton2017inductive} is a general inductive framework that leverages node features to generate node embeddings for previously unseen data. It learns a function that generates embeddings by sampling and aggregating features from a node’s local neighborhood.
    \item \textbf{GAT} \cite{velivckovic2017graph} is a method built on the GCN model. It introduces an attention mechanism at the node level, which allows each node specifies different weights to different nodes in a neighborhood.
    \item \textbf{LGCN} \cite{gao2018large} selects a fixed number of neighboring nodes for each feature based on value ranking in order to transform graph data into grid-like structures. Then, the traditional CNN model is directly applied to learn on the transformed graph.
\end{itemize}

\textit{Methods based on GNN Architecture Search}
\begin{itemize}
    \item \textbf{GraphNAS}  \cite{gao2019graphnas} first uses a recurrent network to generate variable-length strings that describe the architectures of graph neural networks,
    and then trains the recurrent network with reinforcement learning to maximize the embedding accuracy of the generated architectures.
    \item \textbf{Auto-GNN}\cite{zhou2019autognn} is a reinforcement learning-based method similar to GraphNAS, which adopts a parameter sharing strategy that enables homogeneous architectures to share parameters during the training.
\end{itemize}

Following literature \cite{gao2019graphnas}, Chebyshev and GCN are for transductive learning since they require the whole graph structure and nodes to be available in the training. GraphSAGE is used for inductive learning which is able to predict embeddings of unseen graphs based on the trained model. Other baselines including GAT, LGCN, GraphNAS and Auto-GNN are used for both transductive and inductive embedding learning.

\begin{table*}[t]
\renewcommand{\arraystretch}{1.3}
\caption{The transductive node classification results on citation networks.}
\centering
\begin{tabular}{c|c|c|c|c|c}
\toprule
\multirow{2}{*}{\textbf{Categories}} & \multirow{2}{*}{\textbf{Methods}} & \multirow{2}{*}{\textbf{\# Layers}} & \textbf{Cora} & \textbf{Citeseer} & \textbf{Pubmed} \\
\cline{4-6}
\multicolumn{1}{c|}{} & \multicolumn{1}{c|}{} & \multicolumn{1}{c|}{} & \multicolumn{3}{c}{\textbf{Accuracy}} \\
\hline
\multirow{4}{*}{\textbf{Handcrafted GNN Architecture}} & Chebyshev & {$2$} & {$81.2\%$} &   {$69.8\%$} & {$74.4\%$}  \\
& GCN & {$2$} & {$81.5\%$} &   {$70.3\%$} & {$79.0\%$}  \\
& GAT & {$2$} & {$83.0\pm0.7\%$} &   {$72.5\pm0.7\%$} & {$79.0\pm0.3\%$} \\
& LGCN & {$2$} & {$83.3\pm0.5\%$} &   {$73.0\pm0.6\%$} & {$79.5\pm0.2\%$} \\
\hline
\multirow{3}{*}{\textbf{GNN Neural Architecture Search}} & GraphNAS & {$2$} & {$84.2\pm1.0\%$} &   {$73.1\pm0.9\%$} & {$79.6\pm0.4\%$}\\
& Auto-GNN & {$2$} & {$83.6\pm0.3\%$} &   {$73.8\pm0.7\%$} & {$79.7\pm0.4\%$} \\
& \text{\nickname} & {$2$} & {$83.8\pm0.5\%$} &   {$73.5\pm0.8\%$} & {${79.2\pm0.6\%}$} \\
\bottomrule
\end{tabular}
\end{table*}

\subsection{Experimental Setting}
We set the number ($\ns$) of initial structure individuals between 10 and 50, the number ($K_s$) of evolving generations of structure population between 10 and 50, the balance parameter $\alpha$ in Eq. (8) between 0.2 and 1.0. For comparison, the default hyper parameters for \text{\nickname} are set as follows. We set the number of initial structure individuals $\ns$ as 20, the number of initial parameter individuals $\np$ as 6, the number of structure evolving generations $K_s$ as 50, the number of parameter evolving generations $K_p$ as 10, the balance ratio $\alpha$ as 0.6, the numbers of parents for structure and parameter genetics are respectively 10 and 4, the numbers of child for structure and parameter genetics are respectively 4 and 2, the mutation probability for both structure and parameter evolution as 0.02.

For the transductive learning, we aim to identify the optimal architecture of a two-layer GNN model within the search space, while for the inductive learning a three-layer GNN model is optimized in this paper. We train 200 epochs for each specific GNN model, where the accuracy and Micro-F1 are used as metrics for the transductive and inductive embedding learning tasks, respectively.

\begin{table}[t]
\renewcommand{\arraystretch}{1.3}
\caption{The inductive node classification results on the PPI network.}
\centering
\begin{tabular}{c|c|c|c}
\toprule
\multirow{2}{*}{\textbf{Categories}} & \multirow{2}{*}{\textbf{Methods}} & \multirow{2}{*}{\textbf{\# Layers}} & \textbf{PPI} \\
\cline{4-4}
\multicolumn{1}{c|}{} & \multicolumn{1}{c|}{} & \multicolumn{1}{c|}{} & \multicolumn{1}{c}{\textbf{Micro-F1}} \\
\hline
\multirow{4}{*}{\textbf{Handcrafted}} & GraphSAGE (lstm) & {$2$} & {$61.2\%$}  \\
& GAT & {$3$} & {$97.3\pm0.2\%$} \\
& LGCN & {$-$} & {$77.2\pm0.2\%$} \\
\hline
\multirow{3}{*}{\textbf{GNN NAS}} & GraphNAS & {$3$} & {$98.6\pm0.1\%$}\\
& Auto-GNN & {$3$} & {$99.2\pm0.1\%$}\\
& \text{\nickname} & {$3$} & {$98.6\pm0.4\%$}\\
\bottomrule
\end{tabular}
\end{table}
\section{Results}
This section demonstrates the node classification performance of both transductive and inductive graph embedding learning. Then, some important parameters are empirically examined through their impacts on the Cora and Citeseer datasets, respectively. 

\subsection{Graph NAS-based Embedding Learning Performance}
Table V shows the comparative results of all baselines. We can have two main conclusions: 
\begin{itemize}
    \item The NAS-based methods including GraphNAS, Auto-GNN and Genetic-GNN can achieve better results than the handcraft-based GNN models on all three datasets, which verified the effectiveness of NAS to identify good GNN models for the given graph-structure data. This is because the handcrafted models are usually determined by several manual trails, \textit{i.e.}, tuning the number of GNN layers and hyper parameters, which has a very low chance to obtain an optimal model. In comparison, the NAS-based methods are able to search and validate the performance of the candidate GNN models automatically given the graph data and learning task, which can gradually optimize the model performance with litter or even no human intervention.
    \item For the category of NAS-based methods, the performance of our model \text{\nickname} is able to match those of the reinforcement learning-based methods GraphNAS and Auto-GNN. A \textit{t}-student significant test is performed between them with the $p$ value equals to 0.05. It shows that \text{\nickname} is not significantly different, which demonstrate effectiveness of the evolutionary algorithm for GNN architecture search. Compared with the GraphNAS and Auto-GNN, our model is able to optimize both GNN structure and learning parameters, which greatly increased the automation of GNN NAS.
\end{itemize}

\begin{figure}
  \centering
    \includegraphics[width=0.38\textwidth]{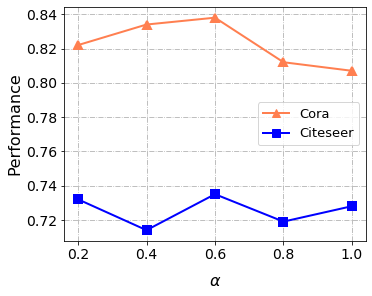}
  \caption{Influence of the imbalance parameter $\alpha$.}%
\end{figure}

Table VI shows the node classification results of inductive embedding learning on the PPI data. Similar conclusions can be made as the transductive learning. First, the category of automated methods perform generally better than the handcrafted methods. Second, performance of our \text{\nickname} model is as good as the state-of-the-art model GraphNAS. In this paper, due to the limitation of computation resource, we only set the maximum population size as 50 and the evolution generations as 50. However, with the increase of population size and evolution generations, our model has a potential to achieve better performance.

\begin{figure}
  \centering
    \includegraphics[width=0.38\textwidth]{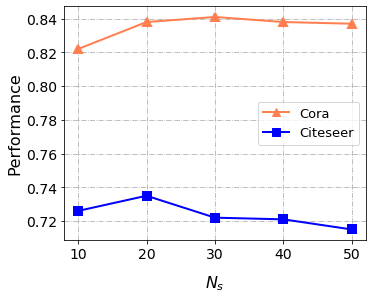}
  \caption{Influence of the number of initial structure population size $K_s$.}%
\end{figure}

\begin{figure}
  \centering
    \includegraphics[width=0.38\textwidth]{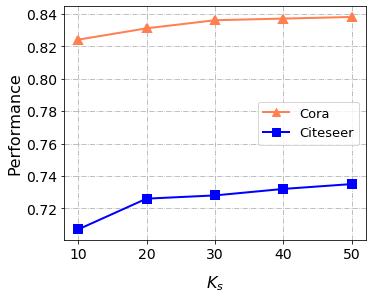}
  \caption{Influence of the number of evolving generations $K_s$ for structure population.}%
\end{figure}

\subsection{Parameter Influence}
We empirically demonstrate the impacts of some important parameters used in \text{\nickname} on Cora and Citeseer data. $\alpha$ is a balance parameter used in the fitness calculation in Eq. (8) and its impact is shown in Fig. 6. We can observe the best setting for both data is 0.6. Fig. 7 shows the influence of the number of structure population size $\ns$. We can observe with the increase of $\ns$, the performance first goes up and then decrease on both citation networks. Generally, larger population size means larger search space which tends to generate better resulting solution. However, the large search space normally requires more evolution generations to finally identify an optimal model, which probably explains the performance decrease as population size increases in Fig. 7. Fig. 8 shows the influence of evolution generation for the structure population, where the performance gradually increases over the generations. 

\begin{figure}
  \centering
    \includegraphics[width=0.38\textwidth]{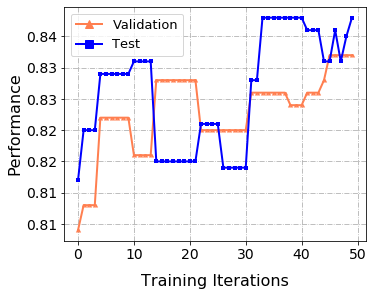}
  \caption{Validation and test performance on Cora change over the training iterations.}%
\end{figure}

\begin{figure}
  \centering
    \includegraphics[width=0.38\textwidth]{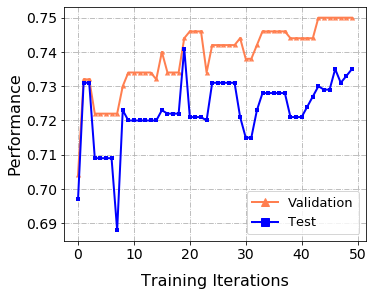}
  \caption{Validation and test performance on Citeseer change over the training iterations.}%
\end{figure}

Fig. 9 and Fig. 10 present the validation and test performances change with the training iterations on Cora and Citeseer, respectively. We can observe the performances have a tendency to improve with the training, through they have some turbulence. The reason for the unstable curves is that both validation and test sets are unseen, and the best performance on the train set cannot guarantee the best performances on the validation and test sets. 

\section{Conclusion} \label{sec:conclusion}
In this paper, we aim to demonstrate the effectiveness of evolutionary neural architecture search for optimizing graph neural network models on graph-structured data. We proposed a novel genetic-based approach called \text{\nickname} for automatically identifying the optimal GNN model with a well-defined search space. Instead of only optimizing the GNN structures with fixed learning parameters, \text{\nickname} is able to evolve and optimize both structure and parameter to fit each other. The experimental results and parameter sensitivity tests demonstrated our model is able to match the state-of-the-art reinforcement learning-based methods.

Since the evolutionary algorithms tend to achieve better solutions with larger population size and evolution generations, it is a future work to test on larger search space and evolution generations. In addition, parameter sharing between individual models is also an interesting direction, \textit{i.e.}, when a parameter individual evolves to a another parameter individual, their model structures remain the same, thereby the model weight parameters can be shared between the two models.

\nocite{*}
\bibliographystyle{IEEEtran}
\bibliography{IEEEabrv,references}

\begin{thebibliography}{10}
\providecommand{\url}[1]{#1}
\csname url@samestyle\endcsname
\providecommand{\newblock}{\relax}
\providecommand{\bibinfo}[2]{#2}
\providecommand{\BIBentrySTDinterwordspacing}{\spaceskip=0pt\relax}
\providecommand{\BIBentryALTinterwordstretchfactor}{4}
\providecommand{\BIBentryALTinterwordspacing}{\spaceskip=\fontdimen2\font plus
\BIBentryALTinterwordstretchfactor\fontdimen3\font minus
  \fontdimen4\font\relax}
\providecommand{\BIBforeignlanguage}[2]{{%
\expandafter\ifx\csname l@#1\endcsname\relax
\typeout{** WARNING: IEEEtran.bst: No hyphenation pattern has been}%
\typeout{** loaded for the language `#1'. Using the pattern for}%
\typeout{** the default language instead.}%
\else
\language=\csname l@#1\endcsname
\fi
#2}}
\providecommand{\BIBdecl}{\relax}
\BIBdecl

\bibitem{zhang2018network}
D.~Zhang, J.~Yin, X.~Zhu, and C.~Zhang, ``Network representation learning: A
  survey,'' \emph{IEEE Transactions on Big Data}, 2018.

\bibitem{cui2018survey}
P.~Cui, X.~Wang, J.~Pei, and W.~Zhu, ``A survey on network embedding,''
  \emph{IEEE Transactions on Knowledge and Data Engineering}, vol.~31, no.~5,
  pp. 833--852, 2018.

\bibitem{wu2020comprehensive}
Z.~Wu, S.~Pan, F.~Chen, G.~Long, C.~Zhang, and S.~Y. Philip, ``A comprehensive
  survey on graph neural networks,'' \emph{IEEE Transactions on Neural Networks
  and Learning Systems}, 2020.

\bibitem{kipf2017semi}
T.~N. Kipf and M.~Welling, ``Semi-supervised classification with graph
  convolutional networks,'' in \emph{International Conference on Learning
  Representations (ICLR)}, 2017.

\bibitem{chen2019multi}
Z.-M. Chen, X.-S. Wei, P.~Wang, and Y.~Guo, ``Multi-label image recognition
  with graph convolutional networks,'' in \emph{Proc. of IEEE CVPR}, 2019, pp.
  5177--5186.

\bibitem{sun2019graph}
M.~Sun, S.~Zhao, C.~Gilvary, O.~Elemento, J.~Zhou, and F.~Wang, ``Graph
  convolutional networks for computational drug development and discovery,''
  \emph{Brief. in bioinformatics}, 2019.

\bibitem{zhao2019t}
L.~Zhao, Y.~Song, C.~Zhang, Y.~Liu, P.~Wang, T.~Lin, M.~Deng, and H.~Li,
  ``T-gcn: A temporal graph convolutional network for traffic prediction,''
  \emph{IEEE Transactions on Intelligent Transportation Systems}, 2019.

\bibitem{zhang2020deep}
Z.~Zhang, P.~Cui, and W.~Zhu, ``Deep learning on graphs: A survey,'' \emph{IEEE
  Transactions on Knowledge and Data Engineering}, 2020.

\bibitem{kriege2020survey}
N.~M. Kriege, F.~D. Johansson, and C.~Morris, ``A survey on graph kernels,''
  \emph{Applied Network Science}, vol.~5, no.~1, pp. 1--42, 2020.

\bibitem{hamilton2017inductive}
W.~Hamilton, Z.~Ying, and J.~Leskovec, ``Inductive representation learning on
  large graphs,'' in \emph{Advances in Neural Information Processing Systems},
  2017, pp. 1024--1034.

\bibitem{zoph2016neural}
B.~Zoph and Q.~V. Le, ``Neural architecture search with reinforcement
  learning,'' \emph{arXiv preprint arXiv:1611.01578}, 2016.

\bibitem{elsken2018neural}
T.~Elsken, J.~H. Metzen, and F.~Hutter, ``Neural architecture search: A
  survey,'' \emph{arXiv preprint arXiv:1808.05377}, 2018.

\bibitem{dong2019one}
X.~Dong and Y.~Yang, ``One-shot neural architecture search via self-evaluated
  template network,'' in \emph{Proceedings of the IEEE International Conference
  on Computer Vision}, 2019, pp. 3681--3690.

\bibitem{chen2019renas}
Y.~Chen, G.~Meng, Q.~Zhang, S.~Xiang, C.~Huang, L.~Mu, and X.~Wang, ``Renas:
  Reinforced evolutionary neural architecture search,'' in \emph{Proceedings of
  the IEEE Conference on computer vision and pattern recognition}, 2019, pp.
  4787--4796.

\bibitem{gao2019graphnas}
Y.~Gao, H.~Yang, P.~Zhang, C.~Zhou, and Y.~Hu, ``Graphnas: Graph neural
  architecture search with reinforcement learning,'' 2019.

\bibitem{zhou2019autognn}
K.~Zhou, Q.~Song, X.~Huang, and X.~Hu, ``Auto-gnn: Neural architecture search
  of graph neural networks,'' 2019.

\bibitem{su2020network}
C.~Su, J.~Tong, Y.~Zhu, P.~Cui, and F.~Wang, ``Network embedding in biomedical
  data science,'' \emph{Briefings in bioinformatics}, vol.~21, no.~1, pp.
  182--197, 2020.

\bibitem{micheli2009neural}
A.~Micheli, ``Neural network for graphs: A contextual constructive approach,''
  \emph{IEEE Transactions on Neural Networks}, vol.~20, no.~3, pp. 498--511,
  2009.

\bibitem{defferrard2016convolutional}
M.~Defferrard, X.~Bresson, and P.~Vandergheynst, ``Convolutional neural
  networks on graphs with fast localized spectral filtering,'' in
  \emph{Advances in neural information processing systems}, 2016, pp.
  3844--3852.

\bibitem{niepert2016learning}
M.~Niepert, M.~Ahmed, and K.~Kutzkov, ``Learning convolutional neural networks
  for graphs,'' in \emph{International conference on machine learning}, 2016,
  pp. 2014--2023.

\bibitem{zhang2016tline}
X.~Zhang, W.~Chen, and H.~Yan, ``Tline: Scalable transductive network
  embedding,'' in \emph{Asia Information Retrieval Symposium}.\hskip 1em plus
  0.5em minus 0.4em\relax Springer, 2016, pp. 98--110.

\bibitem{li2015gated}
Y.~Li, D.~Tarlow, M.~Brockschmidt, and R.~Zemel, ``Gated graph sequence neural
  networks,'' \emph{arXiv preprint arXiv:1511.05493}, 2015.

\bibitem{atwood2016diffusion}
J.~Atwood and D.~Towsley, ``Diffusion-convolutional neural networks,'' in
  \emph{Advances in neural information processing systems}, 2016, pp.
  1993--2001.

\bibitem{velivckovic2017graph}
P.~Veli{\v{c}}kovi{\'c}, G.~Cucurull, A.~Casanova, A.~Romero, P.~Lio, and
  Y.~Bengio, ``Graph attention networks,'' \emph{arXiv preprint
  arXiv:1710.10903}, 2017.

\bibitem{liu2019auto}
C.~Liu, L.-C. Chen, F.~Schroff, H.~Adam, W.~Hua, A.~L. Yuille, and L.~Fei-Fei,
  ``Auto-deeplab: Hierarchical neural architecture search for semantic image
  segmentation,'' in \emph{Proceedings of the IEEE conference on computer
  vision and pattern recognition}, 2019, pp. 82--92.

\bibitem{wang2020textnas}
Y.~Wang, Y.~Yang, Y.~Chen, J.~Bai, C.~Zhang, G.~Su, X.~Kou, Y.~Tong, M.~Yang,
  and L.~Zhou, ``Textnas: A neural architecture search space tailored for text
  representation.'' in \emph{AAAI}, 2020, pp. 9242--9249.

\bibitem{wistuba2019survey}
M.~Wistuba, A.~Rawat, and T.~Pedapati, ``A survey on neural architecture
  search,'' \emph{arXiv preprint arXiv:1905.01392}, 2019.

\bibitem{jaafra2019reinforcement}
Y.~Jaafra, J.~L. Laurent, A.~Deruyver, and M.~S. Naceur, ``Reinforcement
  learning for neural architecture search: A review,'' \emph{Image and Vision
  Computing}, vol.~89, pp. 57--66, 2019.

\bibitem{baker2016designing}
B.~Baker, O.~Gupta, N.~Naik, and R.~Raskar, ``Designing neural network
  architectures using reinforcement learning,'' \emph{arXiv preprint
  arXiv:1611.02167}, 2016.

\bibitem{cai2018efficient}
H.~Cai, T.~Chen, W.~Zhang, Y.~Yu, and J.~Wang, ``Efficient architecture search
  by network transformation,'' in \emph{Thirty-Second AAAI conference on
  artificial intelligence}, 2018.

\bibitem{he2019automl}
X.~He, K.~Zhao, and X.~Chu, ``Automl: A survey of the state-of-the-art,''
  \emph{arXiv preprint arXiv:1908.00709}, 2019.

\bibitem{bergstra2011algorithms}
J.~S. Bergstra, R.~Bardenet, Y.~Bengio, and B.~K{\'e}gl, ``Algorithms for
  hyper-parameter optimization,'' in \emph{Advances in neural information
  processing systems}, 2011, pp. 2546--2554.

\bibitem{hutter2011sequential}
F.~Hutter, H.~H. Hoos, and K.~Leyton-Brown, ``Sequential model-based
  optimization for general algorithm configuration,'' in \emph{International
  conference on learning and intelligent optimization}.\hskip 1em plus 0.5em
  minus 0.4em\relax Springer, 2011, pp. 507--523.

\bibitem{kandasamy2018neural}
K.~Kandasamy, W.~Neiswanger, J.~Schneider, B.~Poczos, and E.~P. Xing, ``Neural
  architecture search with bayesian optimisation and optimal transport,'' in
  \emph{Advances in neural information processing systems}, 2018, pp.
  2016--2025.

\bibitem{ho2020neural}
K.~Ho, A.~Gilbert, H.~Jin, and J.~Collomosse, ``Neural architecture search for
  deep image prior,'' \emph{arXiv preprint arXiv:2001.04776}, 2020.

\bibitem{xie2017genetic}
L.~Xie and A.~Yuille, ``Genetic cnn,'' in \emph{Proceedings of the IEEE
  international conference on computer vision}, 2017, pp. 1379--1388.

\bibitem{suganuma2017genetic}
M.~Suganuma, S.~Shirakawa, and T.~Nagao, ``A genetic programming approach to
  designing convolutional neural network architectures,'' in \emph{Proceedings
  of the genetic and evolutionary computation conference}, 2017, pp. 497--504.

\bibitem{vaswani2017attention}
A.~Vaswani, N.~Shazeer, N.~Parmar, J.~Uszkoreit, L.~Jones, A.~N. Gomez,
  {\L}.~Kaiser, and I.~Polosukhin, ``Attention is all you need,'' in
  \emph{Advances in neural information processing systems}, 2017, pp.
  5998--6008.

\bibitem{sheikh2008genetic}
R.~H. Sheikh, M.~M. Raghuwanshi, and A.~N. Jaiswal, ``Genetic algorithm based
  clustering: a survey,'' in \emph{2008 First International Conference on
  Emerging Trends in Engineering and Technology}.\hskip 1em plus 0.5em minus
  0.4em\relax IEEE, 2008, pp. 314--319.

\bibitem{srivastava2014dropout}
N.~Srivastava, G.~Hinton, A.~Krizhevsky, I.~Sutskever, and R.~Salakhutdinov,
  ``Dropout: a simple way to prevent neural networks from overfitting,''
  \emph{The journal of machine learning research}, vol.~15, no.~1, pp.
  1929--1958, 2014.

\bibitem{gao2018large}
H.~Gao, Z.~Wang, and S.~Ji, ``Large-scale learnable graph convolutional
  networks,'' in \emph{Proceedings of the 24th ACM SIGKDD International
  Conference on Knowledge Discovery \& Data Mining}, 2018, pp. 1416--1424.

\bibitem{shang2018edge}
C.~Shang, Q.~Liu, K.-S. Chen, J.~Sun, J.~Lu, J.~Yi, and J.~Bi, ``Edge
  attention-based multi-relational graph convolutional networks,''
  \emph{arXiv}, pp. arXiv--1802, 2018.

\end{thebibliography}

\end{document}